\documentclass{article}

\usepackage[preprint]{neurips_2026}
\usepackage[utf8]{inputenc} 
\usepackage[T1]{fontenc}    
\usepackage{hyperref}       
\usepackage{url}            
\usepackage{booktabs}       
\usepackage{amsfonts}      
\usepackage{nicefrac}      
\usepackage{microtype}     
\usepackage{xcolor}        
\usepackage{graphicx}     
\usepackage{subcaption}   
\usepackage{float}        
\usepackage{caption}      
\usepackage{wrapfig}
\usepackage{amsmath}
\usepackage{multirow} 
\usepackage{booktabs}
\usepackage{multirow}
\usepackage[ruled,vlined,linesnumbered]{algorithm2e}
\newcommand{\methodname}{\textit{Sword}} 

\title{\textit{Sword}:Style-Robust World Models as Simulators via Dynamic Latent Bootstrapping for VLA Policy Post-Training}

\author{%
  Jiaxuan Gao\thanks{Equal contribution} \\
  Tianjin University \\
\And
  Yongjian Guo\footnotemark[1
  ] \\
  Tsinghua University \\
  JDT AI Infra \\
\And
  Zhong Guan \\
  Tianjin University\\
\And
Wen Huang \\
    Tsinghua University\\
\And
  Wanlun Ma\\
  Swinburne University \\ of Technology\\
\And
  Xi Xiao\\
  Tsinghua University\\
\And
  Junwu Xiong\\
  JDT AI Infra\\
\And
  Sheng Wen\\
  Swinburne University \\ of Technology\\
}

\begin{document}

\maketitle

\begin{abstract}
The integration of Vision-Language-Action (VLA) models with World Models has gained increasing attention. One representative approach treats learned World Models as generative simulators, enabling policy optimization entirely within “imagination.” However, when deployed as simulators for specific environments such as the LIBERO benchmark, existing World Models often suffer from poor generalization and long-horizon error accumulation. During closed-loop rollouts, these models are highly sensitive to initial-state perturbations; minor changes in color, illumination, and other visual factors can trigger cascading hallucinations, leading to severe blurriness or overexposure. Moreover, long-horizon error accumulation further degrades the quality and fidelity of predicted future states. These issues limit the reliability of World Models as simulators. To mitigate these problems, we propose \methodname, a robust World Model framework. Our method introduces Structure-Guided Style Augmentation to disentangle the visual textures of interactive environments from task-relevant dynamics, thereby improving generalization. We further propose Dynamic Latent Bootstrapping, which maintains consistency between training and inference while keeping memory consumption low. Extensive experiments on the LIBERO benchmark show that our method significantly outperforms the baseline WoVR in terms of generalization, generation quality, robustness, fidelity, and the success rate of reinforcement-learning post-training for VLA models.
\end{abstract}

\begin{wrapfigure}{r}{0.5\linewidth}
  \vspace{-0.8em}
  \includegraphics[width=0.98\linewidth]{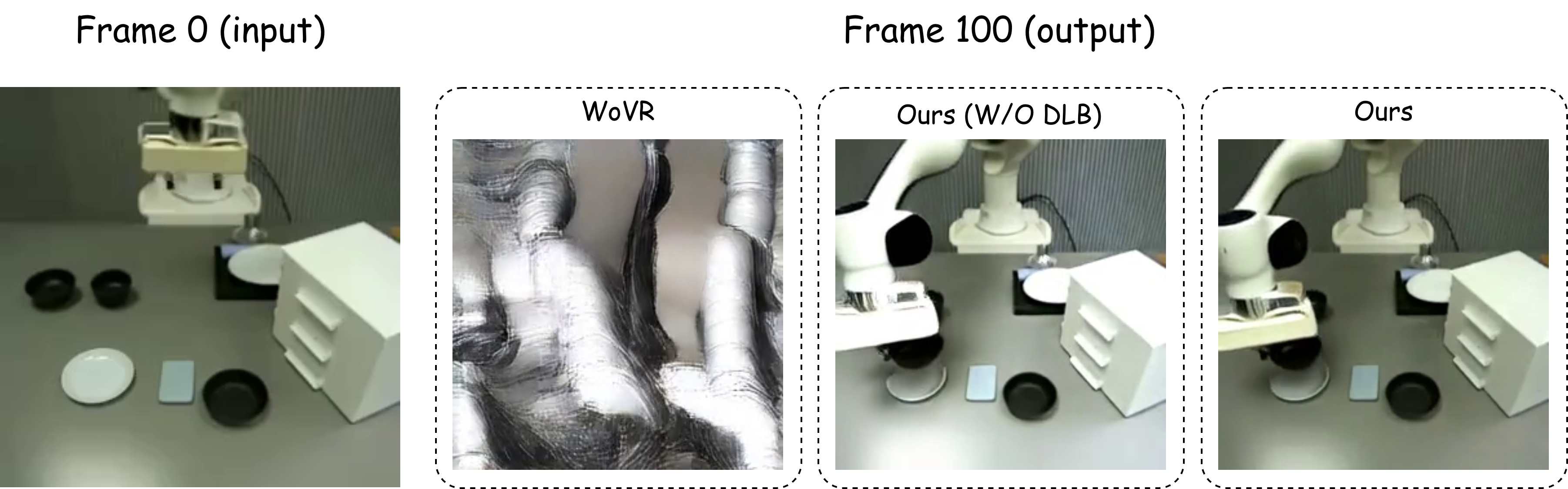}
  \caption{We propose a new world model, Sword, and compare its predicted video frames with those of its variant without Dynamic Latent Bootstrapping (Ours w/o DLB) and WovR~\cite{jiang2026wovr}.}
  \label{fig:teaser}
  \vspace{-0.8em}
\end{wrapfigure}

\section{Introduction}

% --- LaTeX Source Code ---

The development of Vision-Language-Action (VLA) models~\cite{kim2025openvla,bjorck2025gr00t} has marked a critical milestone in robotic manipulation, enabling end-to-end action generation conditioned on multimodal inputs~\cite{ma2024survey,zhong2025survey,zhang2025pure}. While imitation learning has established a strong foundation for these models, the paradigm fundamentally limits the performance ceiling to the quality of human demonstrations. Reinforcement learning (RL) offers a principled mechanism to surpass this bottleneck~\cite{guan2026rl}, yet the prohibitive cost of physical robot interaction necessitates the use of simulated environments. Consequently, World Models have emerged as a compelling alternative, acting as generative simulators that approximate transition dynamics. Recent literature explores the synergistic relationship between these domains, primarily diverging into two paradigms: the first emphasizes the co-evolution of world models and VLA policies within a unified architecture, as demonstrated by frameworks such as RynnVLA~\cite{cen2025rynnvla} and WorldVLA~\cite{cen2025worldvla}, where action understanding and visual generation mutually enhance one another; the second paradigm, exemplified by WoVR~\cite{jiang2026wovr}, treats the world model strictly as a surrogate simulator for post-training policies via reinforcement learning, explicitly regulating imagined rollouts to mitigate hallucination.

Despite these structural advancements, using world models as simulators still exposes two key limitations. 
% The first limitation is insufficient generalization.
The first is an environment-level generalization problem, where the model fails under OOD visual or state perturbations. Current world models are prone to severe overfitting when trained on narrow data distributions, adapting primarily to the specific textures and static configurations of a single environment. When evaluated on the LIBERO benchmark~\cite{liu2023libero}, these models exhibit clear fragility under out-of-distribution (OOD) perturbations in the initial state, such as changes in background color or lighting conditions. Such discrepancies can rapidly trigger catastrophic cascading hallucinations, manifested as severe blurriness or abnormal exposure, thereby depriving the simulator of its ability to model physical dynamics. This failure mode suggests that existing models often merely capture static statistical correlations, rather than learning the underlying dynamic laws. 

The second is an autoregressive rollout problem, where teacher forcing creates a mismatch between ground-truth contexts during training and self-generated contexts during inference.
A world model is typically formulated as a state transition function, taking context frames and actions as input and producing predicted observation frames of future states. However, current world models commonly adopt the Teacher Forcing paradigm~\cite{WMPO2025, jiang2026wovr}, where ground-truth frames are used as context frames during training to predict future observations. In contrast, during inference, the model must condition on its own predicted frames for autoregressive generation. This discrepancy introduces a distribution shift between training and inference, leading to severe \textit{exposure bias}, where small errors in the initial predictions accumulate and are amplified over long-horizon rollouts.

To overcome the above limitations, we propose \methodname, a two-part systematic methodology designed to construct a robust generative simulator by disentangling environmental semantics and aligning the training and inference distributions. To suppress overfitting to superficial visual features, we first introduce a style augmentation method based on the Cosmos~\cite{nvidia2026worldsimulationvideofoundation} architecture. During training, it incorporates diverse perturbations in background appearance, tabletop color, and lighting conditions, encouraging the model to focus more on physical dynamics and geometric constraints. Meanwhile, to eliminate the gap between training and inference, we propose a Dynamic Latent Bootstrapping (DLB) training mechanism. This method continuously leverages the model's own latent predictions as context to reconstruct the forward training process, thereby establishing a conditional distribution consistent with that at inference time. To avoid the storage overhead caused by pixel-space unrolling, we maintain a dynamic cache in the VAE latent space, reducing the memory cost of storing context frames from several hundred GB to under 20 GB and enabling efficient closed-loop optimization. 
The main contributions of this paper are summarized as follows:
\begin{itemize}
    \item \textbf{Structure-Guided Style Augmentation.} 
    We introduce a style augmentation pipeline that diversifies visual conditions while preserving geometric structure and task semantics, encouraging the world model to learn transition-relevant dynamics instead of overfitting to superficial textures, improving its generalization.
    \item \textbf{Dynamic Latent Bootstrapping.} 
    We propose a memory-efficient latent bootstrapping mechanism that gradually conditions the model on its own predicted latents during training, thereby mitigating the train-inference mismatch caused by teacher forcing.
    \item \textbf{Comprehensive Evaluation.} We systematically evaluate Sword against the state-of-the-art WoVR on LIBERO across OOD generalization, prediction quality, simulator fidelity, and ablation studies,  showing its effectiveness as a learned simulator for VLA policy post-training.
\end{itemize}

\section{Related Work}
The intersection of visual generation and robotic control has catalyzed the evolution of Vision-Language-Action models~\cite{zitkovich2023rt2, black2024pi_0, gemini2025robotics, kim2024openvla}. Early frameworks relied predominantly on behavior cloning~\cite{cadene2026lerobot,zhou2026thousand}, mapping visual and linguistic inputs directly to continuous or discrete action spaces. To transcend the limitations of static datasets, on-policy reinforcement learning has been adapted for post-training~\cite{guan2026rl, shukor2025smolvla}, though it is historically bottlenecked by the sample inefficiency of real-world environments. This constraint has motivated the integration of World Models~\cite{wan2025wan,esser2024scaling}, which learn to approximate the environment's transition function $P(o_{t+1}|o_t, a_t)$. Recent architectures like RynnVLA~\cite{cen2025rynnvla} and WorldVLA~\cite{cen2025worldvla} advocate for a unified model where policy derivation and future state prediction share a latent space, demonstrating that action conditioning improves visual fidelity and vice versa. Conversely, frameworks like WoVR~\cite{jiang2026wovr} decouple the components, utilizing an action-conditioned diffusion transformer strictly as a reliable simulator to execute algorithms like PPO~\cite{schulman2017proximal} or GRPO~\cite{shao2024deepseekmath} entirely in imagination.

While these world-model-based simulators demonstrate promise, they inherit the systemic flaws of autoregressive video generation, most notably exposure bias. Traditional training methodologies rely on teacher forcing~\cite{WMPO2025, jiang2026wovr}, evaluating the step-wise loss using ground-truth historical frames. During closed-loop inference, the model must condition on its own imperfect outputs, leading to rapid degradation—a phenomenon exacerbated when the policy explores out-of-distribution states. Recent literature addresses this through techniques like Distribution Matching Distillation~\cite{yin2024one} or Diffusion Forcing~\cite{chen2024diffusion}, which inject varying levels of noise into the context frames to simulate inference conditions. The Self-Forcing paradigm~\cite{huang2025self} further advances this by explicitly unrolling the autoregressive generation during training, forcing the model to learn to recover from its own errors. 
To bridge the distribution shift in long-horizon embodied tasks, we propose Dynamic Latent Bootstrapping (DLB), a mechanism that constructs a "bootstrapped" training loop by recursively feeding generated latents back as context, ensuring the model traverses inference-consistent error pathways to enhance robustness.

\section{Methodology }

\begin{figure}[h!]
    \centering
    \includegraphics[width=1\linewidth]{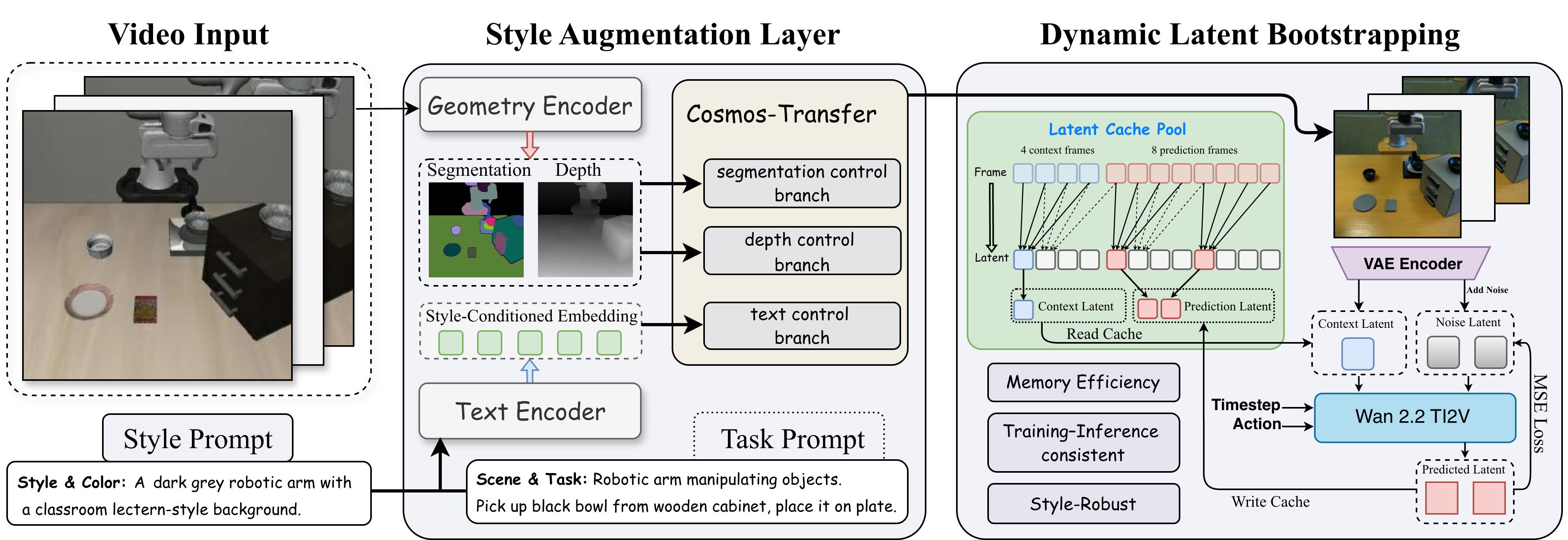}
    \caption{The pipeline of \methodname. %可\jiaxuan{Further details need to be added to align the description of Methodology}
    }
    \label{fig:framework}
\end{figure}

We define the world model as a learned state transition function that predicts the next observation conditioned on the current observation and action:
\begin{equation}
\hat{o}_{t+1} \sim \hat{P}_\phi(o_{t+1} \mid o_t, a_t),
\end{equation}
where $\hat{P}_\phi$ is parameterized by a diffusion-based generative Transformer. By training a reliable world model, we enable the VLA policy to optimize its actions within a learned simulation environment, thereby reducing reliance on real-environment interactions.

\subsection{Structure-Guided Style Augmentation}

To improve data efficiency and prevent the world model from overfitting to low-level visual textures tied to specific environment instances, we introduce a stochastic style augmentation strategy during training without increasing the dataset size. This strategy enhances data reuse and encourages the model to learn invariant physical dynamics from visually diverse yet semantically consistent observations.

Specifically, we construct a style transfer pipeline based on Cosmos-Transfer 2.5~\cite{nvidia2025cosmostransfer1conditionalworldgeneration} and formulate it as a stochastic augmentation function:
\begin{equation}
\tilde{o}_t = \mathcal{A}(o_t, style_t; \eta),
\end{equation}
where $o_t$ denotes the observation, $style_t$ represents the target style specification, and $\eta$ controls the intensity of the style transformation. The function $\mathcal{A}(\cdot)$ applies a series of continuous style transformations to $o_t$, including brightness adjustment, saturation variation, robotic arm color transfer, and modifications to the tabletop and background appearance.

\paragraph{Structure Guidance via Geometric and Task Priors.}

While style augmentation improves visual diversity, unconstrained style transfer may introduce physically implausible artifacts or semantic inconsistencies. To address this issue, we further incorporate structural guidance based on geometric and task priors to stabilize the augmentation process.

We extract auxiliary structural modalities from each observation $o_t$, including depth maps $d_t = \text{Depth}(o_t)$ and semantic segmentation masks $s_t = \text{Seg}(o_t)$, using pretrained encoders such as DepthAnything~\cite{depthanything}, GroundingDINO~\cite{liu2023grounding}, and SAM2~\cite{ravi2024sam2}. These modalities explicitly encode geometric structure and object-level semantics.

To preserve semantic consistency during robot-object interactions, we additionally model the interaction sequence as a task prior $task_t$. This prior is concatenated with the style target $style_t$, forming a joint conditional prompt $p_t = [task_t; style_t]$.

The structural priors $d_t$ and $s_t$ are injected into the geometric control branch of Cosmos-Transfer 2.5 to ensure accurate preservation of key geometric structures during style transfer. Meanwhile, the joint prompt $p_t$ is encoded by a T5-based text encoder~\cite{2020t5}, yielding a textual embedding $e_t = \text{Enc}_{\text{T5}}(p_t)$, which is then fed into the text control branch. This guides the generation process toward both task-relevant semantic features and the desired target style.

Under these structured conditional constraints, the augmentation function can be equivalently expressed as:
\begin{equation}
\tilde{o}_t = \mathcal{A}(o_t \mid d_t, s_t, e_t; \eta).
\end{equation}

Overall, these transformations preserve the underlying interaction semantics, such as object positions and robot-object dynamics, while significantly diversifying visual appearance. As a result, the model is encouraged to focus on stable physical semantics $\mathcal{S}_{\text{env}}$ and geometric constraints $\mathcal{G}_{\text{env}}$, rather than superficial pixel-level cues. This leads to more robust latent representations and substantially improves generalization under OOD conditions.

\subsection{Dynamic Latent Bootstrapping for Mitigating Exposure Bias}

Standard video models typically adopt Teacher Forcing, where ground-truth latents are used as historical context during training. While effective for stabilizing optimization, this strategy introduces a mismatch between training and autoregressive inference.

Traditional world models are commonly trained in a rollout-based manner, where the model recursively feeds its own predictions back as historical context to simulate future trajectories. However, such rollout-based training is not well suited for simulator-oriented world models. Prediction errors can be repeatedly propagated and amplified throughout the rollout process, causing the generated trajectories to gradually deviate from realistic state transitions. This error accumulation makes training unstable and limits the model's ability to function as a reliable simulator over long horizons.

Instead, world models are typically trained using a sampling-based strategy: for each episode, only a short segment of consecutive frames is sampled, e.g., 12 frames with 4 as context and 8 as prediction targets. This formulation better aligns with the objective of learning local state transitions, but makes direct rollout-based training non-trivial.

A straightforward bootstrapping strategy could be to first train the model with Teacher Forcing and then switch to autoregressive or self-conditioned training. However, such staged bootstrapping introduces an abrupt change in the conditioning distribution, making the learning process insufficiently smooth and potentially destabilizing optimization. In contrast, we aim to expose the model to its own predictions in a gradual and continuous manner, so that it can progressively adapt to prediction-based historical context during training.

To address these challenges, we propose \textit{Dynamic Latent Bootstrapping} (DLB), which smoothly increases the use of model-predicted latents as conditioning signals during training while preserving the efficiency and stability of sampling-based optimization.

\paragraph{Dynamic Latent Cache}
The key idea of DLB is to reuse the model's own predictions as bootstrapped conditioning signals. To achieve this efficiently, we introduce a dynamic latent cache $\mathcal{C}$ that stores predicted latent variables $\hat{z}_t$.

Instead of storing data in pixel space, which would require storage comparable to the full dataset, e.g., hundreds of gigabytes, we operate in the compressed VAE latent space, reducing the storage cost by approximately a factor of 60.

During training, predicted latents are continuously written into the cache and updated online, ensuring that the cached distribution remains aligned with the current model. In the early training stage, we primarily use Teacher Forcing for stability while populating the cache. As training progresses, we gradually increase the probability of replacing ground-truth historical latents with cached predictions.

This dynamic bootstrapping process enables a smooth transition from ground-truth conditioning to prediction-based conditioning. As a result, the training distribution becomes better aligned with the inference distribution.

\paragraph{Training Objective}
The DLB objective is defined as:
\begin{equation}
\mathcal{L}_{\text{DLB}}(\theta) =
\mathbb{E}_{z_t, \epsilon, \tau, a_t, \mathcal{C}_t}
\left[
\left\|
\epsilon -
\epsilon_\theta(z_{t, \tau}, \tau, a_t, \mathcal{C}_t)
\right\|_2^2
\right],
\end{equation}
where $\tau$ denotes the diffusion timestep and $\epsilon \sim \mathcal{N}(0, \mathbf{I})$ is Gaussian noise.
By conditioning on bootstrapped latent variables from the dynamic cache, the model learns to correct errors induced by its own predictions. This reduces exposure bias and significantly improves long-horizon stability and consistency.

\subsection{Model Architecture}

We adopt Wan 2.2 TI2V as the backbone of our world model, and inject the action sequence into the model in a textual form to guide future frame prediction. The backbone follows a standard DiT architecture, where actions are incorporated into each Transformer block through two MLP branches via AdaLN and Cross-Attention, respectively. 

At each inference step, the model takes 4 frames as context and predicts the next 8 frames. Since the Wan VAE applies a temporal compression factor of 4, this corresponds to using a single context latent in the latent space to predict two future latents, which are then decoded by the Wan VAE into 8 frames. Furthermore, based on empirical findings from prior work~\cite{shin2026motionstreamrealtimevideogeneration, yang2025longliverealtimeinteractivelong, jiang2026wovr}, we concatenate the first frame of each episode as a global condition with the context at every timestep to enhance temporal consistency across generated frames.

\section{Experiments}
\label{sec:experiments}

In this section, we conduct extensive experiments to evaluate the performance of \methodname. We focus on assessing whether our world model can serve as a high-fidelity simulator for VLA tasks, particularly under challenging OOD conditions.

To comprehensively evaluate the superiority of our world model, we present experimental results from three key perspectives: 
(1) generalization capability, i.e., OOD data (Section~\ref{expr:general}); 
(2) the quality of generated predictions (Section~\ref{expr:qualitative}); 
and (3) the robustness and fidelity of the world model (Section~\ref{expr:robustness}).
% \yongjian{And the libero Success rate?}

\begin{figure}[t]
    \centering
    
    \begin{subfigure}{\linewidth}
        \centering
        \includegraphics[width=\linewidth]{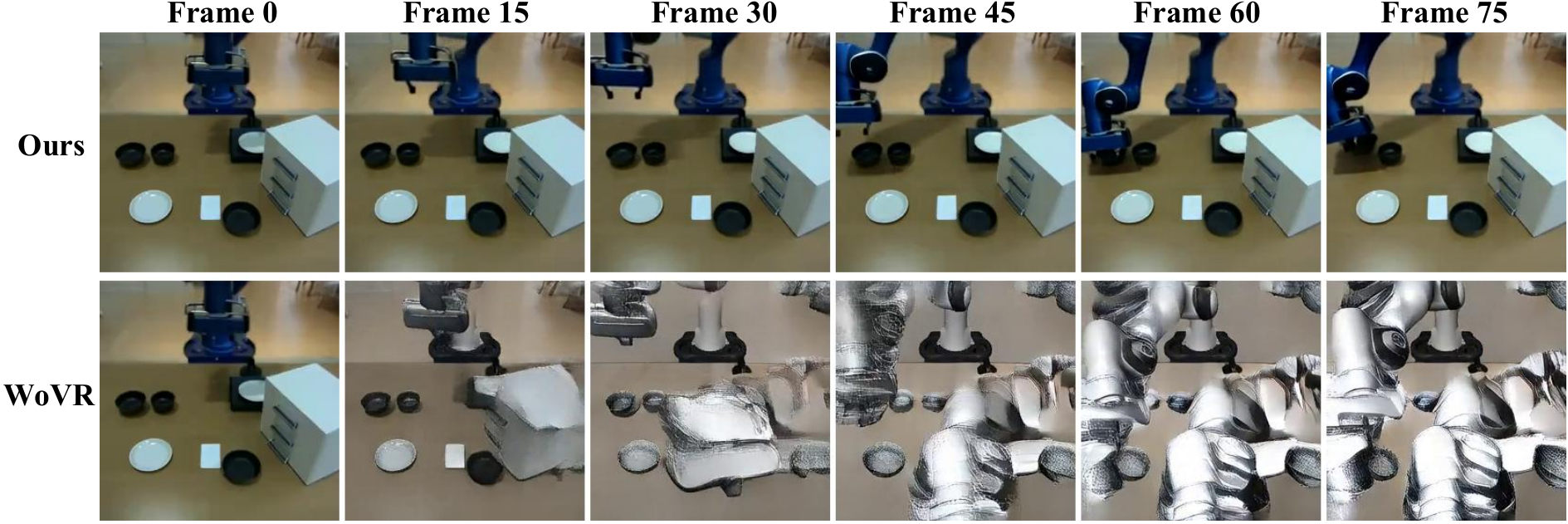}
    \end{subfigure}

    \caption{Qualitative comparison of model performance on OOD data. The baseline (WoVR) fails to generalize, while our model maintains stable and accurate predictions under unseen style variations.}
    \label{fig:generalization}
\end{figure}

\subsection{Experimental Setup}
\label{subsec:setup}

\paragraph{Computational Resources.}
All experiments are primarily conducted on NVIDIA A100 GPUs. We conservatively estimate the total computational cost for training and inference to be approximately 13,000 A100 GPU hours, which corresponds to about 1.5 A100 GPU years.

\paragraph{Dataset and Tasks.}
We conduct experiments on LIBERO-Spatial~\cite{liu2023libero}. We construct a dataset of 1,600 rollout episodes, each containing 512 frames. We use 1,500 episodes for training and the remaining 100 episodes for evaluation. The evaluation set includes \textit{LIBERO-Original}, consisting of raw rollout episodes, and \textit{LIBERO-Mixed}, where half of the episodes are randomly style-augmented to form a 1:1 mixture of raw and OOD data. The augmented samples introduce unseen changes in illumination, saturation, and background textures, and are strictly excluded from training.

\paragraph{Baselines.}
We compare our approach against WoVR~\cite{jiang2026wovr}, a state-of-the-art world model for robotic manipulation simulation proposed in RLinf~\cite{yu2025rlinf}. For a fair comparison, both models use Wan~\cite{wan2025wan} as the backbone and are trained on the same training set for the same number of steps.

\paragraph{Evaluation Metrics.} We employ a suite of quantitative metrics to evaluate generation quality from multiple dimensions:
\textbf{(1) Image Fidelity:} Perceptual similarity and distribution distance are measured using LPIPS~\cite{zhang2018unreasonable} and FID~\cite{heusel2017gans}. \textbf{(2) Video Consistency:} Temporal coherence and video distribution quality are assessed via FVD~\cite{unterthiner2019fvd} and FloLPIPS~\cite{danier2022flolpips}. \textbf{(3) Downstream Utility:} We further report the Success Rate of VLA agents trained/evaluated within the world model to measure its proxy utility for reinforcement learning.

\begin{table}[t]
    \centering
    \begin{minipage}{0.49\textwidth}
        \centering
        \caption{Quantitative comparison of prediction quality.}
        \label{tab:quantitative_results_all}
        \resizebox{\textwidth}{!}{ 
        \begin{tabular}{cccccc}
            \toprule
            Dataset & Model & LPIPS $\downarrow$ & FID $\downarrow$ & FVD $\downarrow$ & FloLPIPS $\downarrow$ \\
            \midrule
            \multirow{2}{*}{\textit{Libero}}
            & WoVR & 0.13 & 22.01 & 61.26 & 0.26 \\
            & Ours & \textbf{0.11} & \textbf{18.39} & \textbf{35.61} & \textbf{0.23} \\
            \midrule
           \textit{Mixed} 
            & WoVR & 0.39 & 119.62 & 198.84 & 0.46 \\
           \textit{(raw + OOD)} & Ours & \textbf{0.20} & \textbf{32.59} & \textbf{111.19} & \textbf{0.30} \\
            \bottomrule
        \end{tabular}
        }
    \end{minipage}
    \hfill 
    \begin{minipage}{0.49\textwidth}
        \centering
        \caption{Quantitative comparison of the full model and the variant without DLB. }
        \label{tab:quantitative_results_abl}
        \resizebox{\textwidth}{!}{
        \begin{tabular}{cccccc}
            \toprule
            Dataset & Model & LPIPS $\downarrow$ & FID $\downarrow$ & FVD $\downarrow$ & FloLPIPS $\downarrow$ \\
            \midrule
            \multirow{2}{*}{\textit{Libero}}
            & w/o DLB  & 0.13 & 20.80 & 48.46 & 0.25 \\
            & Ours & \textbf{0.12} & \textbf{18.51} & \textbf{32.17} & \textbf{0.23} \\
            \midrule
            \multirow{2}{*}{\shortstack{\textit{Mixed} \\ \textit{(raw + OOD)}}}
            & w/o DLB  & 0.22 & 37.84 & 110.71 & 0.32 \\
            & Ours & \textbf{0.17} & \textbf{28.09} & \textbf{86.84} & \textbf{0.26} \\
            \bottomrule
        \end{tabular}
        }
    \end{minipage}
\end{table}

\subsection{Generalization performance Under Distribution Shifts}
\label{expr:general}

We first evaluate the model's performance under OOD conditions. To this end, we construct an OOD test set by applying our style augmentation pipeline to the original test data. This process introduces a diverse range of unseen stylistic variations, including changes in brightness and saturation, robotic arm color transfer, and modifications to the workspace and background. Importantly, all these transformations are strictly excluded from the training data.

We compare \methodname\ against the baseline WoVR~\cite{jiang2026wovr}. As illustrated in Fig.\ref{fig:generalization}, WoVR exhibits a pronounced lack of robustness when encountering distribution shifts. Specifically, starting from Frame 15, its synthesized style begins to deviate from the ground truth (GT, Frame 0); the model fails to accurately fit color saturation and illumination, and geometric artifacts appear in objects such as cabinets. This suggests that WoVR relies on the rote memorization of object appearances rather than a deep understanding of scene semantics, thereby struggling with precise instruction following. By Frame 30, WoVR’s predictions suffer from severe blurring and hallucinations, with object structures completely collapsing—a clear indication of its deficient generalization capabilities. In contrast, our model maintains stable and precise predictions across a variety of unseen styles, consistently preserving stylistic consistency with the initial frame. These results significantly outperform WoVR and provide strong empirical evidence for the superior generalization ability of our proposed \methodname. Additional qualitative results under OOD settings are provided in Appendix Fig.~\ref{fig:ood_result_supp}.

\subsection{Quality of Generated Predictions}
\label{expr:qualitative}

\paragraph{Quantitative Analysis.}

As summarized in Table~\ref{tab:quantitative_results_all}, \methodname\ consistently outperforms WoVR across all metrics. The performance margin is particularly significant on the \textit{LIBERO-Mixed} dataset, where our model achieves a substantial reduction in FVD and FloLPIPS. This indicates that our model effectively mitigates the blurring and jittering common in long-sequence video generation.

\paragraph{Qualitative Comparison.}

\begin{figure}[t]
    \centering
    \begin{subfigure}[b]{0.49\linewidth}
        \centering
        \includegraphics[width=\linewidth]{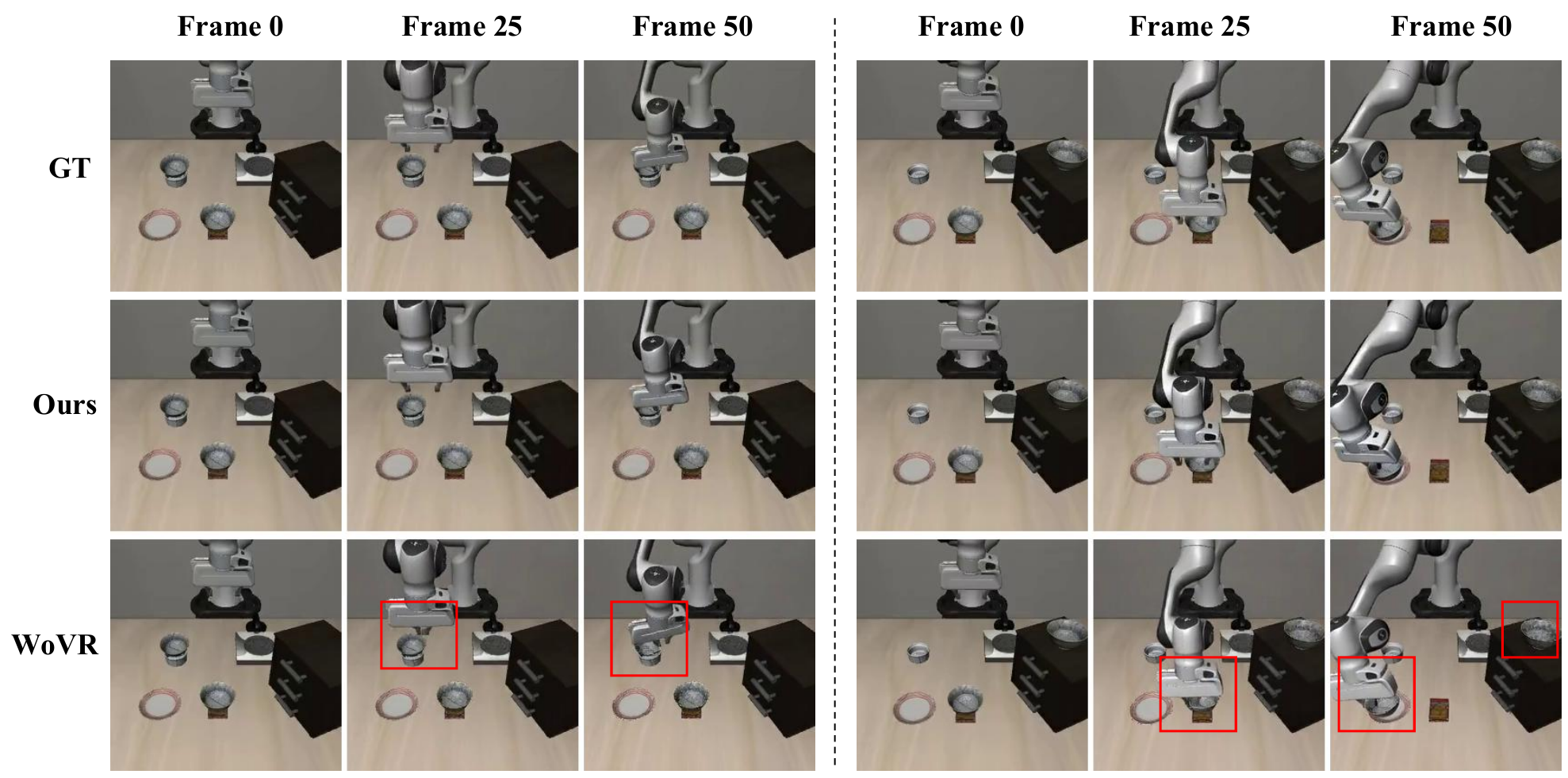}
        \caption{Qualitative comparison of predictions.}
        \label{fig:qualitative}
    \end{subfigure}
    \hfill
    \begin{subfigure}[b]{0.49\linewidth}
        \centering
        \includegraphics[width=\linewidth]{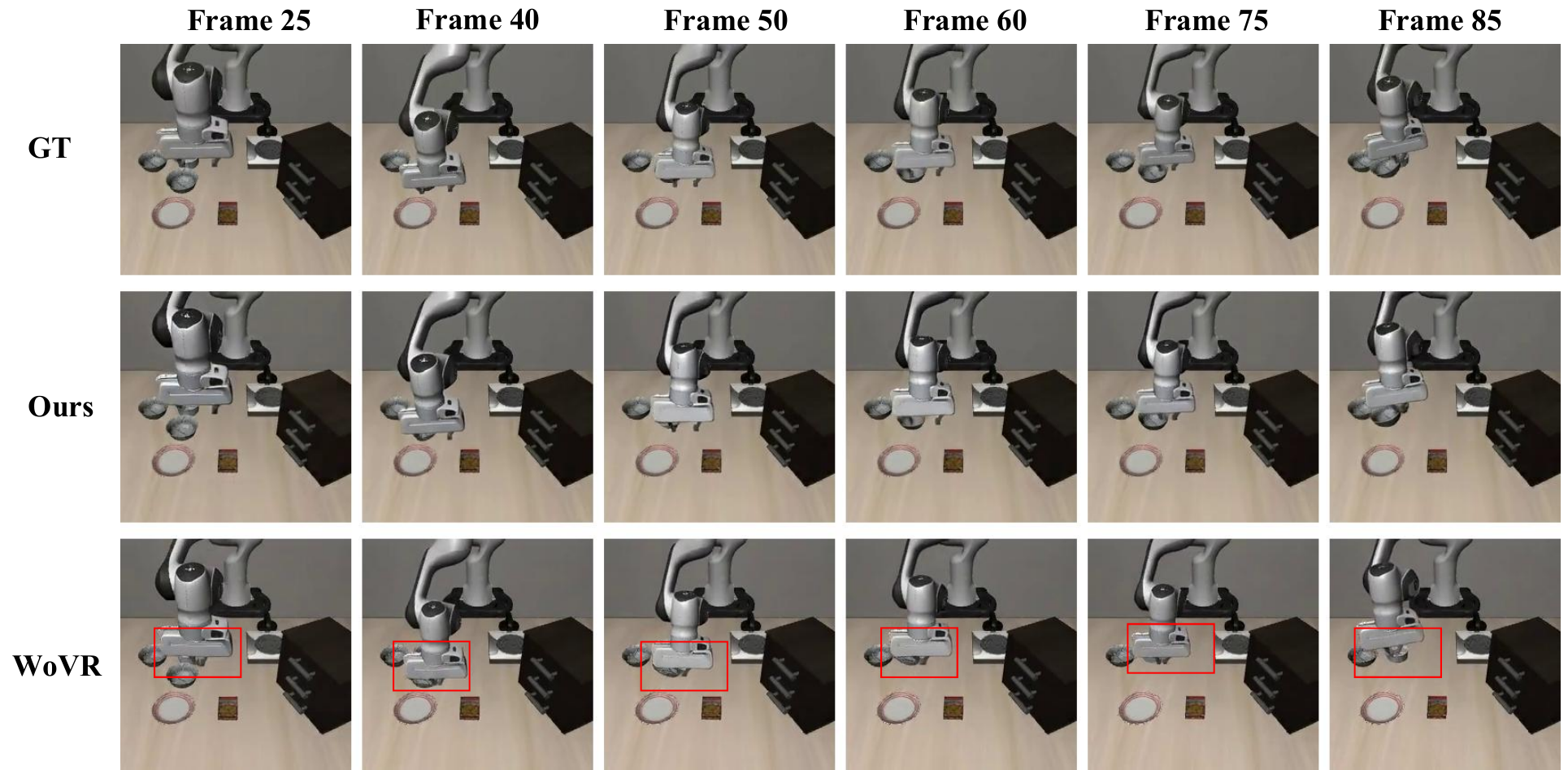}
        \caption{Robustness and fidelity results.}
        \label{fig:robustness}
    \end{subfigure}

    \caption{
        Combined evaluation of our model. 
        \textbf{Left:} Our model produces sharper and more stable long-horizon predictions. 
        \textbf{Right:} Evaluation of action-following ability and physical interaction consistency.
    }
    \label{fig:qualitative_robustness}
\end{figure}

We also perform qualitative evaluation by randomly selecting samples from the original Libero dataset and comparing predicted frames generated by our model and WoVR. As shown in the top part of Fig.~\ref{fig:qualitative_robustness}, \methodname\ produces predictions that are significantly more aligned with the ground truth. Specifically, our world model demonstrates a robust capability to mitigate compounding errors during long-horizon generation. In contrast, WoVR suffers from severe performance degradation over time, characterized by pronounced blurring, noise, and structural distortions. For instance, in the left example at Frame 25, WoVR's gripper closes prematurely before reaching the object, and subsequently re-opens at Frame 50—a clear manifestation of temporal hallucinations. In the right example, the accumulated noise in WoVR's output becomes increasingly dominant, leading to a substantial loss of visual clarity. Conversely, our model yields consistently sharp, stable, and semantically coherent results, underscoring its superior long-term generative quality.

\subsection{Robustness and Fidelity of the World Model}
\label{expr:robustness}

The ultimate goal of a world model is to replace traditional simulators and enable reinforcement learning post-training for VLA agents. Therefore, robustness and fidelity are critical evaluation criteria.
We evaluate both aspects on the original Libero dataset. Given an initial frame and the corresponding action sequence, we assess: 
(1) the model's ability to follow action inputs over long horizons; and 
(2) its ability to maintain semantic consistency in physical interactions.

The results are shown in the right part of Fig.~\ref{fig:qualitative_robustness}. In this example, the ground-truth sequence contains a challenging interaction process: the gripper first approaches the object according to the action inputs, the initial grasp attempt fails and slightly displaces the object, and the agent then follows subsequent corrective actions to complete a successful second grasp.

WoVR struggles to faithfully follow the given action sequence in the early stage. As shown in frames 25, 40, and 50, it prematurely closes the gripper before the intended grasping moment, indicating a mismatch between the generated gripper dynamics and the action inputs. As these errors accumulate over time, the generated sequence diverges from the ground-truth interaction process and eventually produces a semantically different outcome. In particular, after the initial failed grasp, WoVR fails to maintain temporal coherence and effectively ``forgets'' the object in later frames, resulting in a breakdown of interaction semantics.

In contrast, our model accurately follows the action inputs throughout the sequence. The gripper remains open when the robotic arm has not yet approached and aligned with the object. When the arm first moves close to the object but still has a certain positional offset, our model does not blindly close the gripper. Instead, it keeps the gripper open, waits for the arm to adjust to a more accurate grasping position, and then closes the gripper to grasp the object. These results demonstrate that our world model achieves stronger action controllability and better semantic fidelity in long-horizon physical interactions. More qualitative comparisons regarding robustness and fidelity can be found in Appendix Fig.~\ref{fig:physical_fidelity_supp}.

\subsection{Policy Success Rate}

To verify whether our world model can improve the performance of VLA policies, we conduct GRPO reinforcement learning with OpenVLA-OFT on the LIBERO-Spatial benchmark and compare our method with the baseline WoVR. As shown in Table~\ref{tab:policy_success_rate}, our method consistently achieves higher policy success rates across different training steps, demonstrating that the proposed world model provides more effective training feedback for policy optimization.

\section{Ablation Study}
\subsection{Effectiveness of the DLB}

To validate the effectiveness of the proposed \textbf{Dynamic Latent Bootstrapping} (DLB), we conduct ablation studies on both the original Libero dataset and the mixed dataset containing raw and style-transferred OOD samples. We evaluate model performance using LPIPS, FID, FVD, and FloLPIPS.

We compare our full model with a variant where DLB is disabled, i.e., predicted latents stored in the dynamic latent cache are not used as bootstrapped historical conditioning signals during training. As shown in Table~\ref{tab:quantitative_results_abl}, removing DLB consistently degrades performance across all metrics and datasets, with the performance gap remaining clear under the more challenging mixed setting with OOD style shifts. These results demonstrate that DLB improves not only perceptual quality but also long-horizon temporal stability, especially when the model is exposed to distribution shifts.

\begin{figure}[t]
\centering
\includegraphics[width=\linewidth]{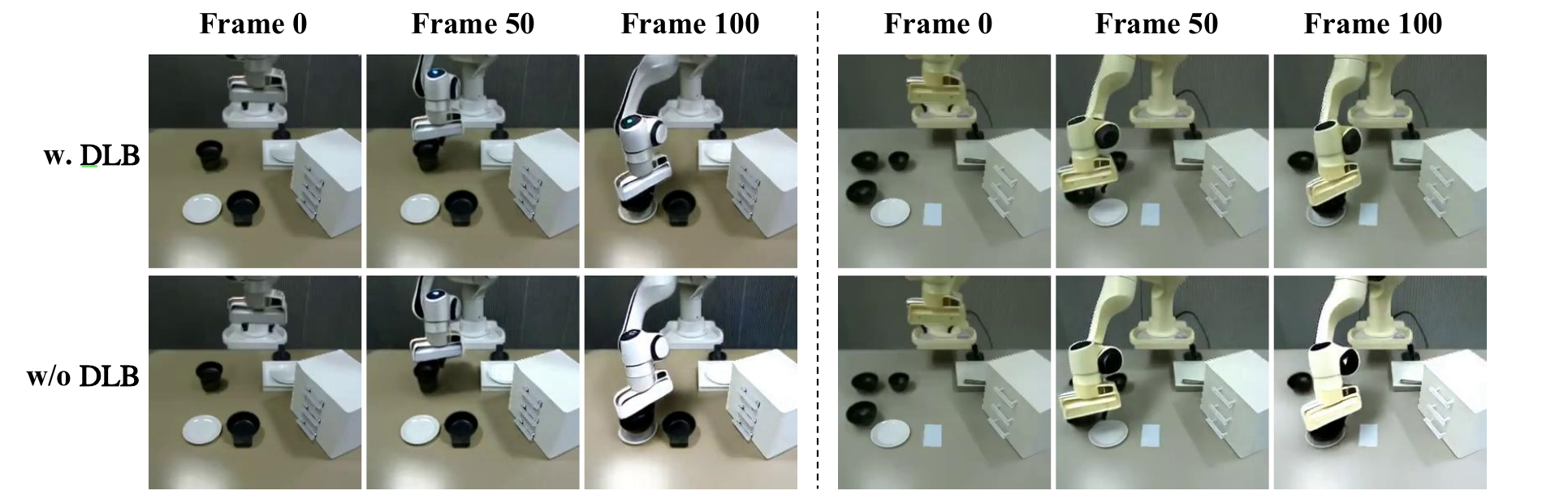}
\caption{Qualitative comparison between the full model and the variant without Dynamic Latent Bootstrapping. Without DLB, the prediction quality deteriorates significantly in later timesteps, exhibiting noticeable inconsistencies in lighting and brightness.}
\label{fig:dlb_qualitative}
\end{figure}

As illustrated in Figure~\ref{fig:dlb_qualitative}, disabling DLB leads to clear degradation in prediction quality during later stages of generation. In particular, the model exhibits severe inconsistencies in illumination and brightness over time. This suggests that conditioning on bootstrapped latent predictions from the dynamic latent cache is important for reducing exposure bias and maintaining temporal coherence during autoregressive generation.

\begin{table}[t]
    \centering
    \caption{Policy success rate comparison at different training steps.}
    \label{tab:policy_success_rate}
    \begin{tabular}{lccccccc}
        \toprule
        Model & Step 10 & Step 35 & Step 65 & Step 95 & Step 120 & Step 150 & Step 172 \\
        \midrule
        WoVR & 41.02\% & 42.58\% & 45.70\% & 39.84\% & 49.22\% & 40.62\% & 53.12\% \\
        Ours & \textbf{43.36\%} & \textbf{48.44\%} & \textbf{49.22\%} & \textbf{50.78\%} & \textbf{58.59\%} & \textbf{60.55\%} & \textbf{61.72\%} \\
        \bottomrule
    \end{tabular}
\end{table}

\subsection{Comparison of Different Cache Design Strategies}

To better understand how different cache designs affect model performance, we conduct a full-factorial ablation study over two major aspects of the cache mechanism: the \emph{read-cache schedule} and the \emph{write-cache design}. The read-cache schedule controls how the model reads from the cache during training, where we compare linear decay with warmup + cosine decay. The write-cache design determines how the cache is updated after each prediction, and consists of two factors: the replacement strategy, i.e., direct replacement vs. EMA replacement, and the write-back scope, i.e., single-chunk write-back vs. all-chunk write-back.

By exhaustively enumerating all possible combinations of these factors, we obtain a total of $2 \times 2 \times 2 = 8$ experimental settings. To highlight the relative performance of different designs, we report the results in Table~\ref{tab:cache_design_fullfactorial}, sorted by FVD in ascending order. From Table~\ref{tab:cache_design_fullfactorial}, we observe that the read-cache schedule has a more noticeable impact on performance. When the results are sorted by FVD, linear decay generally achieves better results than warmup + cosine decay, with the top two configurations both adopting linear decay. In contrast, the influence of different write-cache designs is less regular and less pronounced. Neither direct replacement nor EMA replacement consistently outperforms the other, and the advantage of single-chunk or all-chunk write-back also varies across different read-cache schedules.

\begin{table*}[t]
\centering
\caption{Full-factorial comparison of different cache design strategies. We evaluate two read-cache schedules and four write-cache designs. Results are sorted by FVD in ascending order.}
\label{tab:cache_design_fullfactorial}
\resizebox{\textwidth}{!}{
\begin{tabular}{c c c c|cccc}
\toprule
\textbf{Rank} & \textbf{Read-cache Schedule} 
& \textbf{Replacement Strategy} & \textbf{Write-back Scope}
& \textbf{LPIPS} $\downarrow$ & \textbf{FID} $\downarrow$ & \textbf{FVD} $\downarrow$ & \textbf{FloLPIPS} $\downarrow$ \\
\midrule
1 & Linear Decay            & Direct Replacement & All Chunks   & \textbf{0.17} & \textbf{28.09} & \textbf{86.84} & \textbf{0.26} \\
2 & Linear Decay            & EMA Replacement    & Single Chunk & 0.19 & 30.16 & 86.97 & 0.28 \\
3 & Warmup + Cosine Decay   & Direct Replacement & Single Chunk & 0.19 & 29.25 & 88.44 & 0.28 \\
4 & Linear Decay            & EMA Replacement    & All Chunks   & 0.20 & 31.31 & 89.51 & 0.30 \\
5 & Warmup + Cosine Decay   & EMA Replacement    & All Chunks   & 0.18 & 29.49 & 92.01 & 0.28 \\
6 & Linear Decay            & Direct Replacement & Single Chunk & 0.20 & 34.25 & 95.98 & 0.29 \\
7 & Warmup + Cosine Decay   & Direct Replacement & All Chunks   & 0.20 & 33.89 & 101.20 & 0.29 \\
8 & Warmup + Cosine Decay   & EMA Replacement    & Single Chunk & 0.21 & 35.06 & 103.69 & 0.31 \\
\bottomrule
\end{tabular}
}
\end{table*}

This phenomenon is reasonable because the read-cache schedule directly controls how frequently cached predicted latents are used as historical conditioning signals during training. Therefore, it has a strong influence on the conditioning distribution seen by the model. By contrast, cache write-back is performed after every prediction. As training progresses, especially in the middle and late stages, different write-cache strategies may gradually lead to similar cache-pool distributions due to frequent updates, making their final effects less distinguishable.
Overall, the best performance is achieved by combining \textbf{Linear Decay}, \textbf{Direct Replacement}, and \textbf{All Chunks}, indicating that this configuration provides the most effective cache design in our setting.

\section{Conclusion and Limitations}
\label{sec:conclusion_discussion}
This paper addresses the critical vulnerabilities of using world models as simulators for Vision-Language-Action policy optimization. We identified that current autoregressive simulators suffer from severe overfitting and exposure bias, leading to catastrophic hallucinations when faced with environmental perturbations such as lighting changes or shifted object positions. By introducing style-transfer augmentation and auxiliary structural guidance via depth and segmentation maps, we successfully constrained the model to learn invariant physical dynamics rather than superficial textures. Crucially, we resolved the train-inference gap through a novel latent self-forcing mechanism, utilizing a highly memory-efficient dynamic cache that allows the model to train on its own predicted context. Extensive evaluations on the LIBERO benchmark confirm that our unified approach eliminates the cascading errors prevalent in existing frameworks like WoVR, resulting in a highly robust generative simulator that significantly improves the generalization and overall success rate of the optimized policies in unseen environments. Due to limited computational resources, our experiments are mainly compared against WoVR~\cite{jiang2026wovr}; in future work, we plan to conduct broader comparisons with more world-model-based methods and further explore VLA post-training in style-transferred environments to better validate and improve our approach.

\bibliography{mybib}
\bibliographystyle{unsrt}

%%%%%%%%%%%%%%%%%%%%%%%%%%%%%%%%%%%%%%%%%%%%%%%%%%%%%%%%%%%%
\clearpage

\appendix

\section{Technical appendices and supplementary material}

\subsection{Additional Qualitative Results for Generalization under Distribution Shifts}
\label{app:generalization_ood}

This section provides supplementary qualitative results for Sec.~\ref{expr:general}. These additional comparisons further demonstrate the generalization performance of Sword (Ours) under OOD settings.

\begin{figure}[H]
    \centering
    \includegraphics[width=\linewidth]{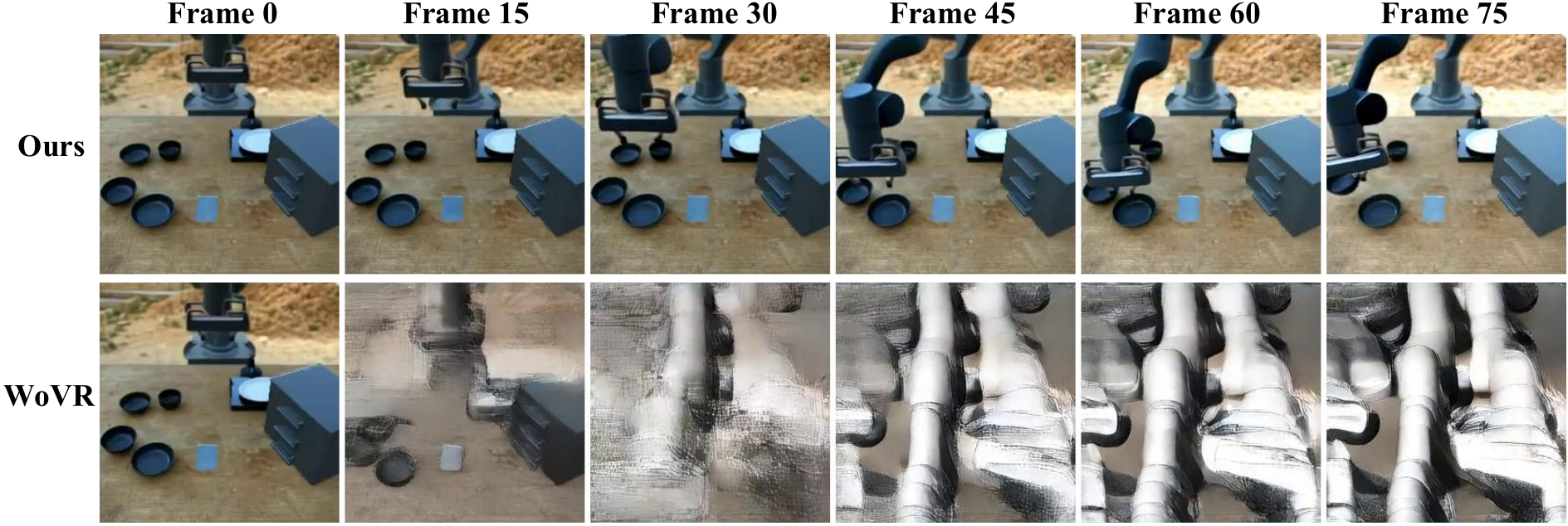}
    \vspace{0.1em}

    \includegraphics[width=\linewidth]{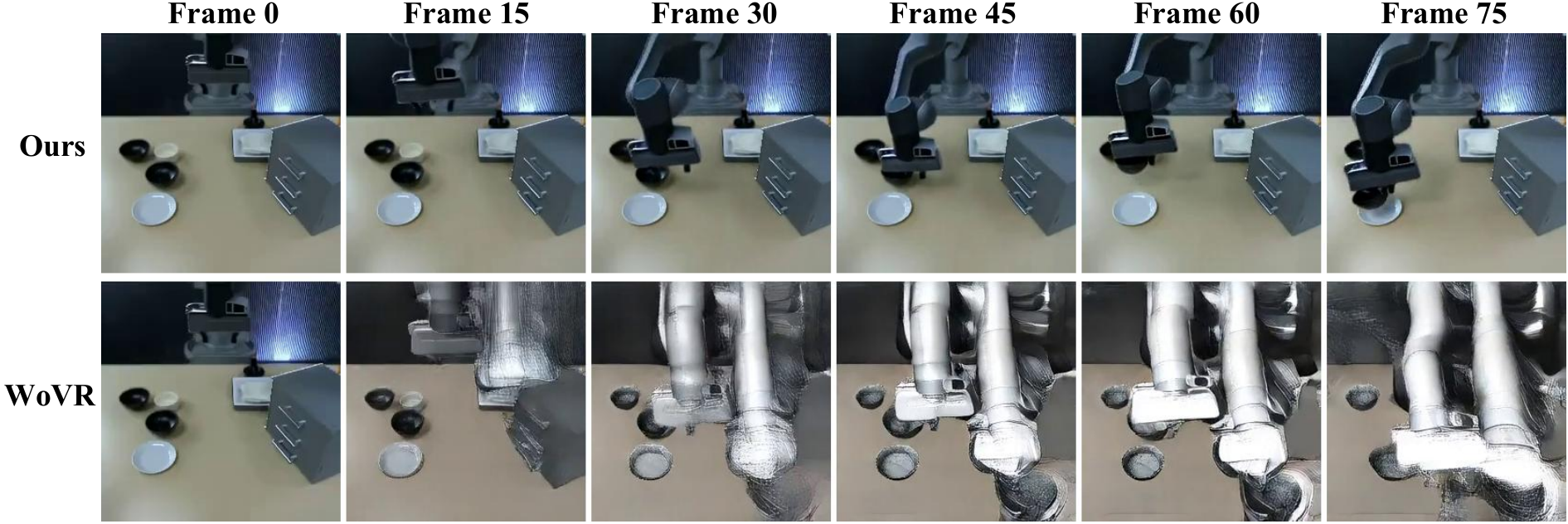}
    \caption{
    \textbf{Additional qualitative comparisons under OOD settings.}
    As a supplement to Sec.~\ref{expr:general}, we provide additional qualitative comparisons between Sword (Ours) and WoVR on OOD data. Sword produces more accurate and temporally consistent predicted frames, demonstrating stronger robustness under distribution shifts.
    }
    \label{fig:ood_result_supp}
\end{figure}

\subsection{Additional Qualitative Results for Physical Fidelity}
\label{app:physical_fidelity}
\begin{figure*}[t]
    \centering
    \includegraphics[width=\textwidth]{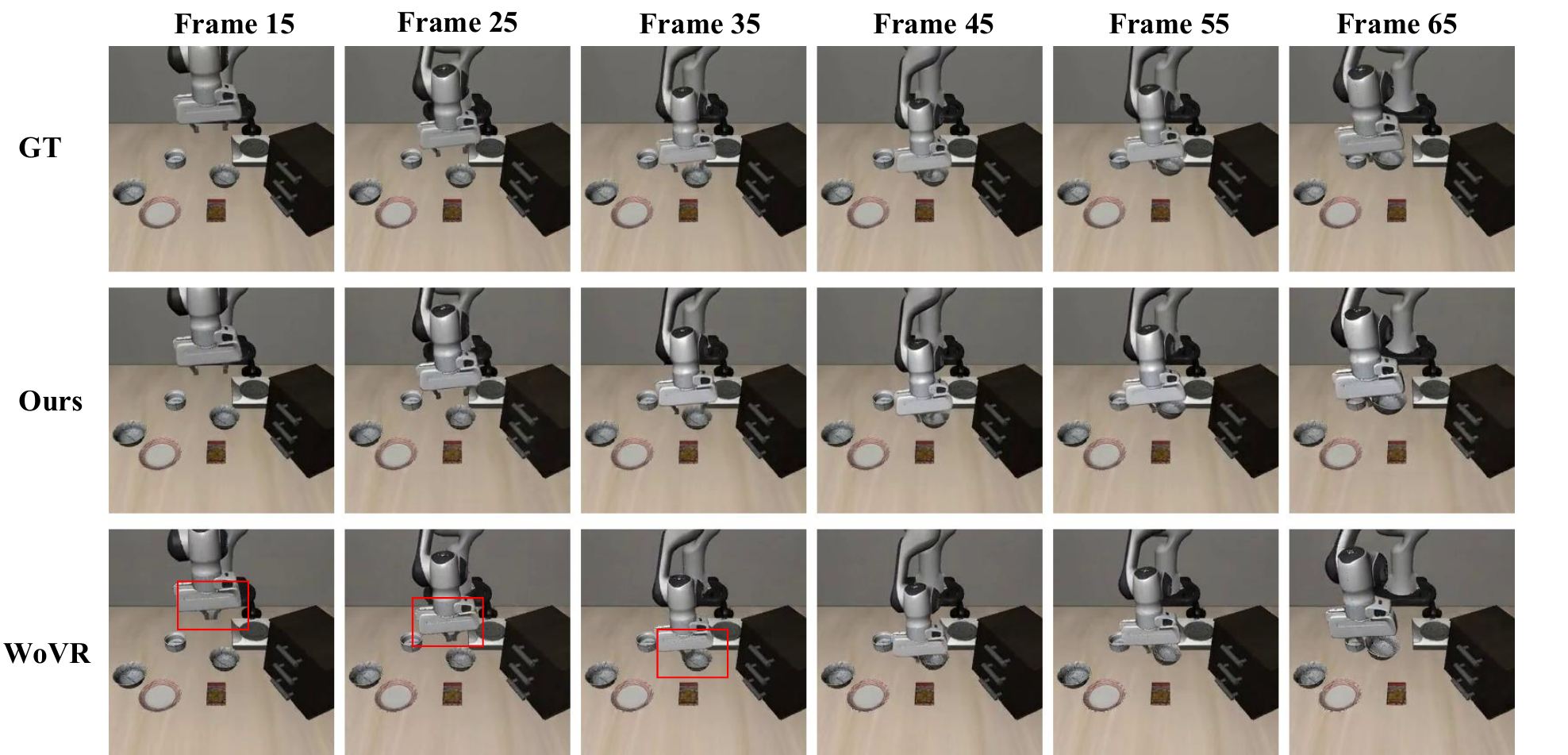}
    \caption{
    \textbf{Additional qualitative comparisons of robustness and fidelity.}
    We provide supplementary results to compare the robustness and fidelity of predicted video frames. Sword (Ours) better follows the input actions, whereas WoVR produces incorrect gripper states of the robotic arm.
    }
    \label{fig:physical_fidelity_supp}
\end{figure*}

% \label{expr:robustness}
This section provides supplementary qualitative results for Sec.~\ref{expr:robustness}. As shown in Fig.~\ref{fig:physical_fidelity_supp}, Sword (Ours) better preserves object states, robot-object interactions, and physically plausible temporal dynamics, further demonstrating its ability to model accurate physical evolution during long-horizon prediction.

%%%%%%%%%%%%%%%%%%%%%%%%%%%%%%%%%%%%%%%%%%%%%%%%%%%%%%%%%%%%

\end{document}